\documentclass[preprint,12pt]{elsarticle}

\usepackage{amssymb}
\usepackage{amsmath}
\usepackage{graphicx}
\usepackage[colorlinks=true, allcolors=blue]{hyperref}
\usepackage{array}
\usepackage{subcaption}
\usepackage{tabularx}

\journal{Acta Astronautica}

\begin{document}

\begin{frontmatter}


%% use the tnoteref command within extcitle for footnotes;

\title{Onboard-Targeted Segmentation of Straylight in Space Camera Sensors}

\author[label1,label2]{Riccardo Gallon\corref{cor1}} %% Author name
\ead{r.gallon@tudelft.nl}
\cortext[cor1]{Corresponding author.}

\author[label2]{Fabian Schiemenz}
% \author[inst2]{Alisa \snm{Krstova}}
\author[label1]{Alessandra Menicucci}
\author[label1]{Eberhard Gill}

\affiliation[label1]{organization={Department of Space Systems Engineering, Faculty of Aerospace Engineering},%Department and Organization
            addressline={Kluyverweg 1}, 
            city={HS Delft},
            postcode={2629},
            country={Netherlands}}

\affiliation[label2]{organization={Airbus Defence and Space GmbH},%Department and Organization
            addressline={Claude-Dornier Stra\ss e}, 
            city={Immenstaad am Bodensee},
            postcode={88090}, 
            country={Germany}}

%% Abstract
\begin{abstract}
This study details an artificial intelligence (AI)-based methodology for the semantic segmentation of space camera faults. Specifically, we address the segmentation of straylight effects induced by solar presence around the camera's Field of View (FoV). Anomalous images are sourced from our published dataset. Our approach emphasizes generalization across diverse flare textures, leveraging pre-training on a public dataset (\textit{Flare7k++}) including flares in various non-space contexts to mitigate the scarcity of realistic space-specific data. A DeepLabV3 model with MobileNetV3 backbone performs the segmentation task. The model design targets deployment in spacecraft resource-constrained hardware. Finally, based on a proposed interface between our model and the onboard navigation pipeline, we develop custom metrics to assess the model's performance in the system-level context.
\end{abstract}


% \begin{highlights}
% \end{highlights}

%% Keywords

\end{frontmatter}



\section{Introduction}

As modern space missions increasingly employ camera sensors for navigation purposes, tailored Fault Detection, Isolation and Recovery (FDIR) approaches are required. Concurrently, spacecraft autonomy requirements become more stringent as space missions are oriented towards scenarios where the human intervention is prevented by the enormous distances and irregular and sparse spacing of contact windows, e.g., planetary exploration and landing. In this framework, FDIR solutions relying on AI can bridge traditional gaps of FDIR based on ECSS-E-ST-70-41C (Packet Utilization Standard, \cite{ECSS-E-ST-70-41C}), enhancing autonomy while enabling multi-channel signal handling (i.e., images), potentially in real-time. This research is driven by the seek for such a solution, focusing on the fault detection task and providing an embedded-compatibility.\\
We consider the \textit{Astrone KI} mission \cite{martin2023pioneering} as use case for providing a suitable scenario to perform the AI-based FDIR task. It consists in a concept vehicle for autonomous relocations on the 67P/Churyumov-Gerasimenko comet, leveraging AI to augment the Guidance \cite{olucak2023sensor}, Navigation \cite{liesch2023ai} and FDIR \cite{gallon2025convolutional, gallon2024machine} subsystems. The sensors employed for attitude and pose estimation include LiDARs and cameras, gyroscopes and accelerometers. As we focus on the camera, the first challenge for the FDIR task is identifying and clustering failure cases for this sensor. These cases are use-case specific, and may be different or inapplicable to other mission scenarios. Limited resources are available in the state of the art about faults occurring in space cameras, despite the consistent presence of these sensors onboard interplanetary exploration missions \cite{bos2020flight, yamada2023inflight, west2010flight, porco2004cassini, soderblom2008mars, bayard2019vision, yingst2020dust}. Typically, straylight effects in space cameras are prevented by baffles mounted on the camera lens to obscure parasitic side light, but they present reliability flaws \cite{bos2020flight}. Software-level solutions do not exist for real-time applications due to the computational burden required. However, the straylight occurrence demands real-time analysis, because of the high variability of this effect in successive acquired frames. AI can provide a fast and reliable way of detecting straylight, improving mission safety and availability.

We elaborated on space camera faults in our previous work \cite{gallon2025addressing}, which provides a list of possible failure modes and a simulation strategy to render resulting artifacts on synthetic images. We also provide an off-the-shelf dataset for AI-based FDIR training and testing \cite{gallon2025addressing}. In the present work, we employ our dataset to perform AI-based semantic segmentation of faults. The criticality of this detection task resides in the fact that fault-induced artifacts in camera imagery can obstruct visibility and lead to erroneous sensor reading. When processed in the onboard software, these corrupted data can degrade performance and/or cause faults in nominal operations. Our objective is to achieve fail-operational conditions where AI accurately segments faulty pixels, enabling their exclusion from subsequent image processing. This prevents the propagation of erroneous data and maintains system functionality. This specific FDIR framework represents a safety-critical application demanding high reliability and rigorous analysis of its integration with other spacecraft components. The integration is indeed fundamental to establish the performance and the reliability of such an AI-based FDIR architecture, and it shall aim to produce performance indices and define requirements which can be extended and reused for the application to future missions.

Our contribution is an onboard-compatible deep learning solution for the semantic segmentation of Straylight, specifically flares and glares originated from sunlight. We evaluate the developed model using novel, system-aware metrics designed considering the processing of the output in downstream onboard algorithms. The selected AI model is onboard-compatible, meaning that it is designed considering resource-constrained deployment. We address the faults segmentation task initially by targeting all faulty classes present in the dataset \cite{gallon2025addressing}, subsequently narrowing down our focus to the Straylight class only. During the design process, challenges associated with the scarcity of data were encountered and efficiently addressed through pre-training over a large existing dataset \cite{dai2023flare7k++}. Finally, considering the interface between the model and the onboard software, we derive custom metrics to measure the performance of the model in a system-level wide scope.

This paper is structured as follows: Section \ref{sec:rel_work} surveys relevant prior research in embedded-compatible semantic segmentation, with a specific focus on its applications in space, alongside a review of related work in straylight analysis. Section \ref{sec:methodology} details the development process of our AI-based segmentation solution for camera faults. Section \ref{sec:ev_res} presents the evaluation metrics and discusses onboard model integration considerations. Finally, Section \ref{sec:concl_futwork} offers conclusions and outlines future work.

\section{Related Work} \label{sec:rel_work}

Semantic segmentation, which involves dense pixel-level predictions, has seen significant advancements through fully convolutional architectures like FCN, U-Net, DeepLab, and GoogLeNet \cite{long2015fully, ronneberger2015u, he2016deep, szegedy2015going}. More recently, Transformers have emerged as a powerful alternative to convolutions, demonstrating strong performance across various tasks, including segmentation \cite{vaswani2017attention, liu2021swin, dosovitskiy2021an}, and notably in Natural Language Processing.\\
A critical challenge has been deploying segmentation models on computationally constrained hardware such as CPUs and mobile devices \cite{chen2024review}. To address this, depthwise separable convolutions, initially seen in Inception and Xception networks \cite{szegedy2015going, chollet2017xception}, became central to the MobileNet series for efficient inference on mobile and embedded systems \cite{howard2017mobilenets, sandler2018mobilenetv2, howard2019searching}. The Inverted Residual Block, introduced in MobileNetV2, significantly improved memory efficiency through residual connections between bottleneck layers \cite{sandler2018mobilenetv2}. Further efficiency gains were achieved by the EfficientNet family, which employed Neural Architecture Search (NAS) to generate models with fewer parameters and faster inference, with EfficientNetV2 specifically utilizing the Fused-MBConv block, a refined Inverted Residual Block \cite{tan2019efficientnet, tan2021efficientnetv2}. Additionally, Dilated (Atrous) Convolutions have been adopted to expand the receptive field without increasing computational or memory overhead, finding use in compact models like MobileNetV3 and ESPNet \cite{yu2015multi, howard2019searching, mehta2018espnet}. For semantic segmentation specifically, Spatial Pyramid Pooling (SPP) is often used for multi-scale context aggregation \cite{he2015spatial}. The main advantages of SPP are two-fold: on the one hand, the concatenation of feature maps obtained by different downsampling rates allows learning features of the image at different scales. In our specific application, this feature helps in segmenting objects possessing ver different sizes. On the other hand, SPP also allows to obtain a fixed-size feature representation, removing any requirement on the size of the network input image. This characteristic is particularly useful to enable data augmentation strategies via cropping the image to supply to data lacking, but it can also be exploited after deployment to process data at different resolutions, i.e., coming from different onboard units. A significant advancement in the field of pyramid pooling is constituted by the DeepLab series \cite{chen2017rethinking, chen2018encoder}, which introduced the Atrous Spatial Pyramid Pooling (ASPP) module. ASPP employs atrous convolutions and global average pooling to provide multi-scale feature learning while maintaining low computational overhead. Low-dimensional features from the backbone are processed in the ASPP module, while bilinear upsampling is utilized to reshape the output to the desired size. In DeepLabV3+ \cite{chen2018encoder}, this upsampling strategy was revised to improve small-scale features learning, as it is identified as the weak point of previous DeepLab architectures \cite{chen2017rethinking}. In \cite{chen2018encoder}, the authors introduce a skip connection from upper layers of the backbone, adequately reshaped, to the output of the ASPP. This strategy facilitates small-objects segmentation by providing context from layers whose downsampling factor is small. \\
In the present work, the deployment in space of the presented compact AI applications is targeted. While these algorithms are typically benchmarked against standard datasets \cite{deng2009imagenet,lin2014microsoft,everingham2010pascal,cordts2016cityscapes} and consumer hardware, we evaluate the model on a space-related dataset and target the deployment on space-grade Commercial Off-The-Shelf (COTS) components. However, space-specific data are acknowledgedly limited, making pre-training on large-scale natural image datasets and subsequent transfer learning crucial for achieving robust performance. The problem of data lacking is particularly critical in the fault detection task faced in this work. To the best of the author's knowledge, very little work exists in the field. The authors of \cite{wang2021unified} propose an AI-based solution for general anomaly detection in remote sensing images, focusing on failure modes occurring in Charged-coupled Devices (CCD) camera sensors which overlap with our dataset \cite{gallon2025addressing}, but not including flares. However, public datasets including straylight effects on quotidian terrestrial scenarios exist \cite{dai2022flare7k,dai2023flare7k++,wu2021train}, and they can be used to supply to our lack of data, i.e., they can be used for pre-training our model, as they provide a statistically significant population of images. \\
Lastly, we reviewed works that addressed the topic of AI compatibility with space-qualified hardware. Space-qualified hardware for AI processing in space is presented in \cite{furano2020towards}. In \cite{bahl2019low}, \textit{lightweight} versions of state-of-the-art AI algorithms (U-Net, FCN) targeting onboard spacecraft deployment are proposed, but the actual deployment process is not detailed. The authors of \cite{schwarz2025early} present an interesting demonstration flying a GPU in space and achieving onboard training. Similarly, the authors of \cite{feresin2021space} present their onboard deployment of a Field-Programmable Gate Array (FPGA) running AI for cloud segmentation on OPS-SAT. Lastly, the work of \cite{precht2024performance} provides hardware benchmarks for three AI models deployed on FPGAs. The models come from typical space-domain use cases, ship detection, hazard detection and avoidance for asteroid landing \cite{caroselli2022neo}, and anomaly detection \cite{gallon2025convolutional}. The anomaly detection use case comes from our previous work and proves the deployability and real-time capability of the model. The hardware used in \cite{precht2024performance} is two different space-grade boards, \textit{Zynq UltraScale+} and \textit{Versal}. \\

\section{Methodology} \label{sec:methodology}
% Dataset (from my previous paper) and Algorithm description\\
% Hardware compatibility, inference speed table comparison with DeepLabV3+ and UNet (at least) $\rightarrow$ justifies \textbf{\textit{onboard}} in the \textbf{\textit{Problem}} statement\\
% Metrics development based on navigation interface $\rightarrow$ descends from \textbf{\textit{integrated at subsystem level}} in the \textbf{\textit{Problem}} statement\\
% Results, metrics evaluation over a representative dataset. It will descend from the original one in my paper, augmented to cover different onboard image processing cases (TBC)

\subsection{Model Search}
The main focus of the model selection process is to design a space hardware-compatible model with reduced memory footprint and Floating-Point Operations (FLOPs) in order to target onboard deployment, i.e., comply with limited available resources. To this aim, we adopt a \textit{DeepLabV3} architecture \cite{chen2017rethinking} with a \textit{MobileNetV3 Large} backbone \cite{howard2019searching}. The $output\_stride$ is set to 16 and the atrous rates in the backbone are kept to (6,12,18),  as suggested in the optimal results obtained by \cite{chen2017rethinking}.\\ 
Due to the transient nature of straylight effect, in this application the real-time processing constraint is more restrictive than other faults which may occur in space cameras (e.g., Broken Pixels, Lines or Dust on Optics). The ability to process image data concurrently with straylight occurrences is crucial, as flares can appear and fade rapidly, potentially affecting system dynamics. In contrast, other faults from \cite{gallon2025addressing} generally have slower dynamics. Dust shows major effects during take-off, and slightly changes in other flight phases, while Broken Pixels and Lines occur once during flight and persist until a proper recovery is applied. This foundational difference enables flexibility in the detection of these faults, where potential inference of subsequent images would improve detection reliability. In addition, live inference constraints can be relaxed and heavier models can be deployed. Conversely straylight detection shall occur in the arc of a single image, in order to be able to evaluate and counteract its effects. This real-time constraint demands higher model performances on single image samples and precludes the processing of image sequences to not exceed hardware resources.

\subsection{Development Process}\label{subs:dev_proc}

\paragraph{Dataset}This study utilizes a proprietary dataset \cite{gallon2025addressing} for the task of fault segmentation. The dataset categorizes faults into five distinct classes: \textit{Dust on Optics}, \textit{Broken Pixels and Lines}, \textit{Straylight}, \textit{Optics Degradation}, and \textit{Vignetting}. While segmentation analysis is primarily conducted on the first three classes, \textit{Optics Degradation} (Blurriness) and \textit{Vignetting} are represented by binary indices rather than spatial masks, and are treated as global disturbance sources. This strategy serves to robustly evaluate our solution's ability to identify faults even within images affected by the mentioned disturbances. Future research is recommended to explore methodologies for detecting and quantifying Blurriness and Vignetting effects, alongside the other aforementioned faults.\\
The dataset employed in this work retains only the fraction of the dataset presented in \cite{gallon2025addressing} that comprises straylight, mainly prevent class imbalance between the straylight class and other faults/nominal pixels. Dataset features are presented in Table \ref{tab:dataset}.
\begin{table}[h!]
    \centering
    \begin{tabular}{|c|c|}
    \hline
    Property & Value \\
    \hline
    Number of Images & 1000 \\    
    \hline
    Training/Validation Split & 80/20 \\
    \hline
    Input Size & $1024\times1024\times3$ \\
    \hline
    Color Bit Depth & 8 bits per channel \\
    \hline
    Number of Classes & 2 (\textit{Nominal}, \textit{Straylight}) \\    
    \hline
    \end{tabular}
    \caption{Custom dataset properties}
    \label{tab:dataset}
\end{table}

\begin{figure}[h!]
    \centering
    \includegraphics[width=0.6\linewidth]{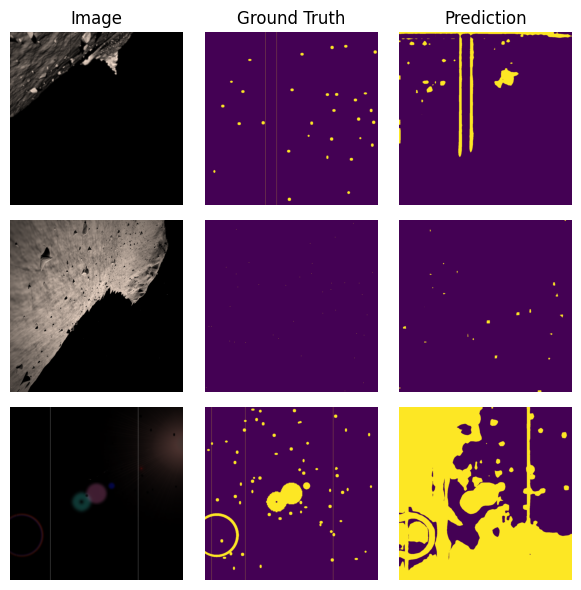}
    \caption{Binary segmentation of all dataset fault classes with DeepLabV3. Training data is described in Table \ref{tab:dataset}. Training run for 500 epochs with Binary Cross-entropy loss and Adam optimizer with $0.0001$ initial learning rate.}
    \label{fig:multiclass_allfaults}
\end{figure}

\paragraph{Faults Clustering}We conducted first experiments by training the model on the full dataset, performing binary segmentation of faults where annotation is available (i.e., Dust on Optics, Broken Pixels and Lines, Straylight). Results from the DeepLabV3 model \cite{chen2017rethinking} showed good performances on the flares (Figure \ref{fig:multiclass_allfaults}). 

The segmentation masks for Straylight reveals a clear understanding of the position and shape of individual textures. Nevertheless, the specific asteroid environment and the texture injection methodology used to simulate flares can introduce biases in the data. Such an issue cannot be evaluated for other faults of our dataset due to the substantial absence of relevant public datasets, but for the straylight case it is possible to leverage public datasets to obtain better performances from the model in recognising unseen textures (dataset modelling limitation) on unseen backgrounds (generalization capability).  

\begin{figure}[h!]
    \centering
    % Row 1
    \begin{subfigure}[t]{0.24\textwidth}
        \centering
        \includegraphics[width=\textwidth]{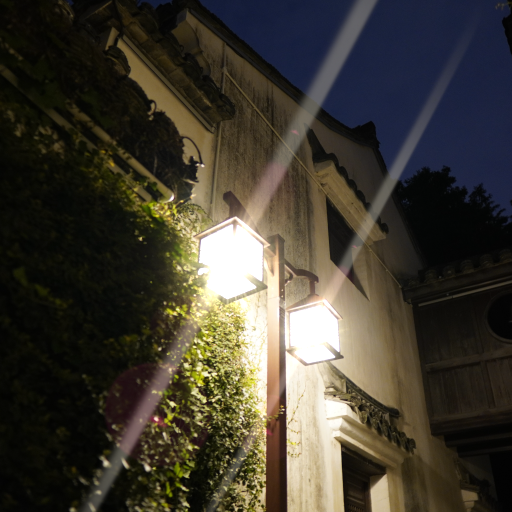}
    \end{subfigure}
    \hfill
    \begin{subfigure}[t]{0.24\textwidth}
        \centering
        \includegraphics[width=\textwidth]{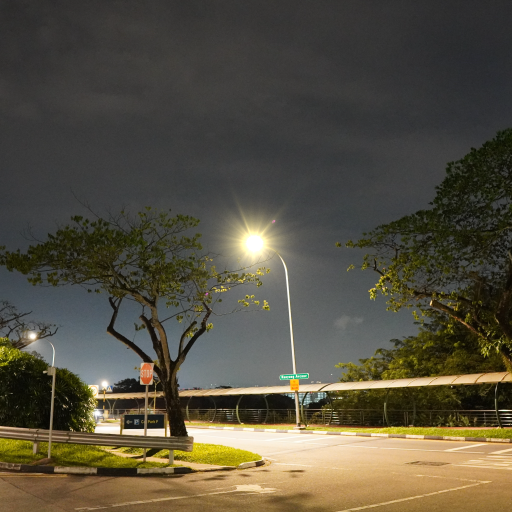}
    \end{subfigure}
    \hfill
    \begin{subfigure}[t]{0.24\textwidth}
        \centering
        \includegraphics[width=\textwidth]{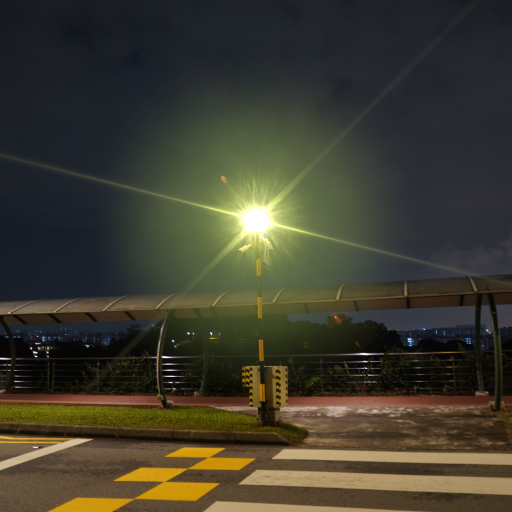}
    \end{subfigure}
    \hfill
    \begin{subfigure}[t]{0.24\textwidth}
        \centering
        \includegraphics[width=\textwidth]{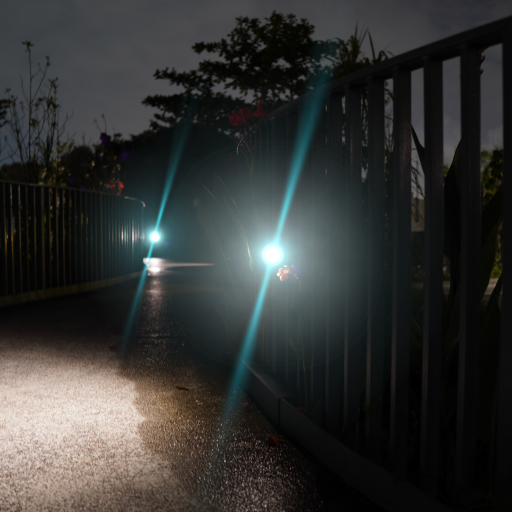}
    \end{subfigure}

    \vspace{0.1cm} % New line with a small vertical space for the next row

    % Row 2
    \begin{subfigure}[t]{0.24\textwidth}
        \centering
        \includegraphics[width=\textwidth]{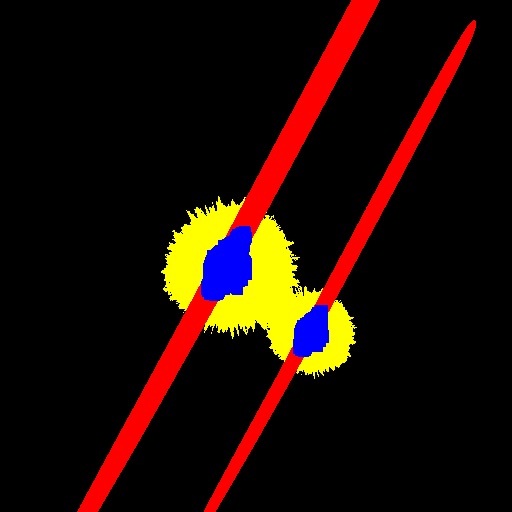}
    \end{subfigure}
    \hfill
    \begin{subfigure}[t]{0.24\textwidth}
        \centering
        \includegraphics[width=\textwidth]{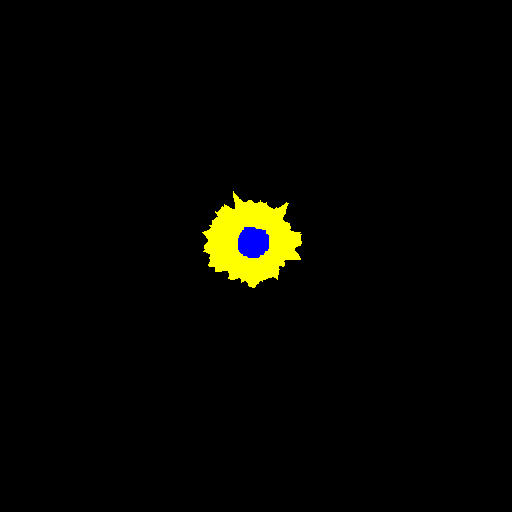}
    \end{subfigure}
    \hfill
    \begin{subfigure}[t]{0.24\textwidth}
        \centering
        \includegraphics[width=\textwidth]{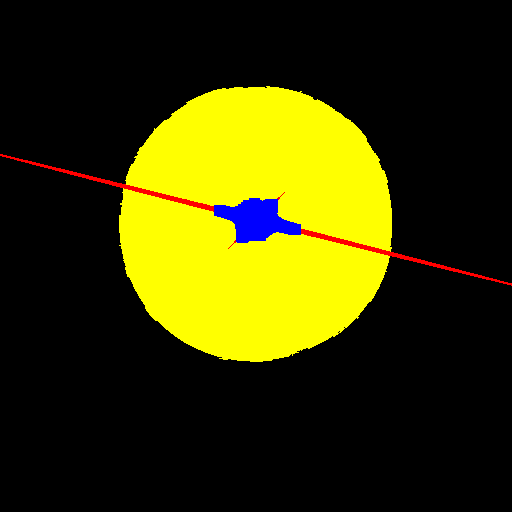}
    \end{subfigure}
    \hfill
    \begin{subfigure}[t]{0.24\textwidth}
        \centering
        \includegraphics[width=\textwidth]{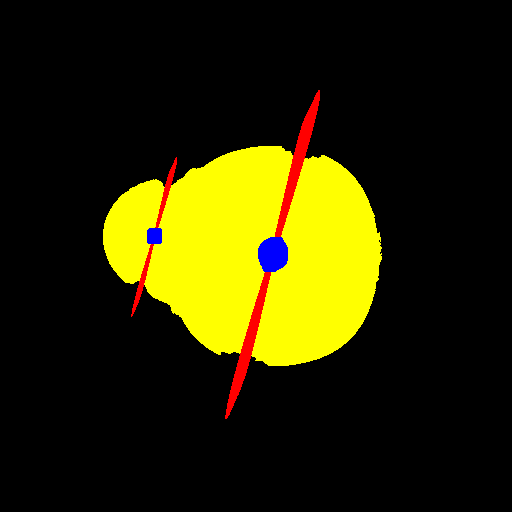}
    \end{subfigure}
    
    \caption{Sample images synthetically generated and relative segmentation mask from Flare7k++ \cite{dai2023flare7k++}}
    \label{fig:flare7kpp}
\end{figure}

\paragraph{Final Model Design}The target of the analysis shall be two-fold. First, our model shall aim to robustly segment unseen flare textures that could appear in real-world data. The dataset exposed in \cite{gallon2025addressing} does not provide realistic flares modelling because it is not in scope, but it is still possible to achieve good generalization of the trained model by leveraging pre-training on large flares datasets. Second, large dataset pre-training also helps the model in generalizing over different backgrounds. Our dataset is indeed space-specific, but generic background learning could have beneficial effects on reusing the model in different mission scenarios (e.g., planetary exploration). We employ the Flare7k++ dataset \cite{dai2022flare7k, dai2023flare7k++} for pre-training our model. Dataset samples and specifications are given in Figure \ref{fig:flare7kpp} and Table \ref{tab:flare7kpp_specs}. We pre-trained for 100 epochs using binary cross-entropy loss before fine-tuning on our proprietary dataset from \cite{gallon2025addressing}. This process allows the model to learn more generalizable features of flares across diverse image backgrounds, thereby enhancing its robustness to various textures and scenarios.
\begin{table}[h!]
    \centering
    \begin{tabular}{|c|c|}
    \hline
    Property & Value \\
    \hline
    Number of Images & 7962 \\
    \hline
    Input Size & $512\times512\times3$ \\
    \hline
    Color Bit Depth & 8 bits per channel \\
    \hline
    Number of Classes & 2 (\textit{Nominal}, \textit{Straylight}) \\    
    \hline
    \end{tabular}
    \caption{Flare7k++ dataset properties \cite{dai2023flare7k++}}
    \label{tab:flare7kpp_specs}
\end{table}

\begin{figure}[h!]
    \centering
    \includegraphics[width=0.6\textwidth]{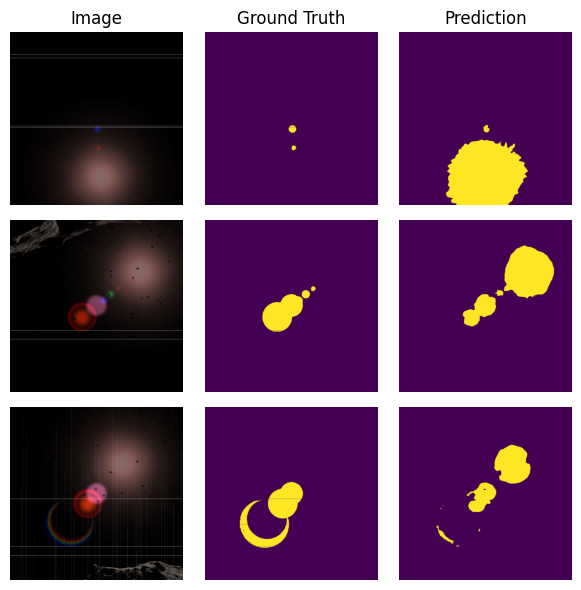}
    \caption{Binary segmentation of straylight. Pre-trained model inferenced on custom test dataset}
    \label{fig:binclass_sl}
\end{figure}

To validate the efficacy of pre-training for the flare segmentation task, we conducted an initial test on our custom dataset, as depicted in Figure \ref{fig:binclass_sl}.

\begin{figure}[h!]
    \centering
    \includegraphics[width=0.6\textwidth]{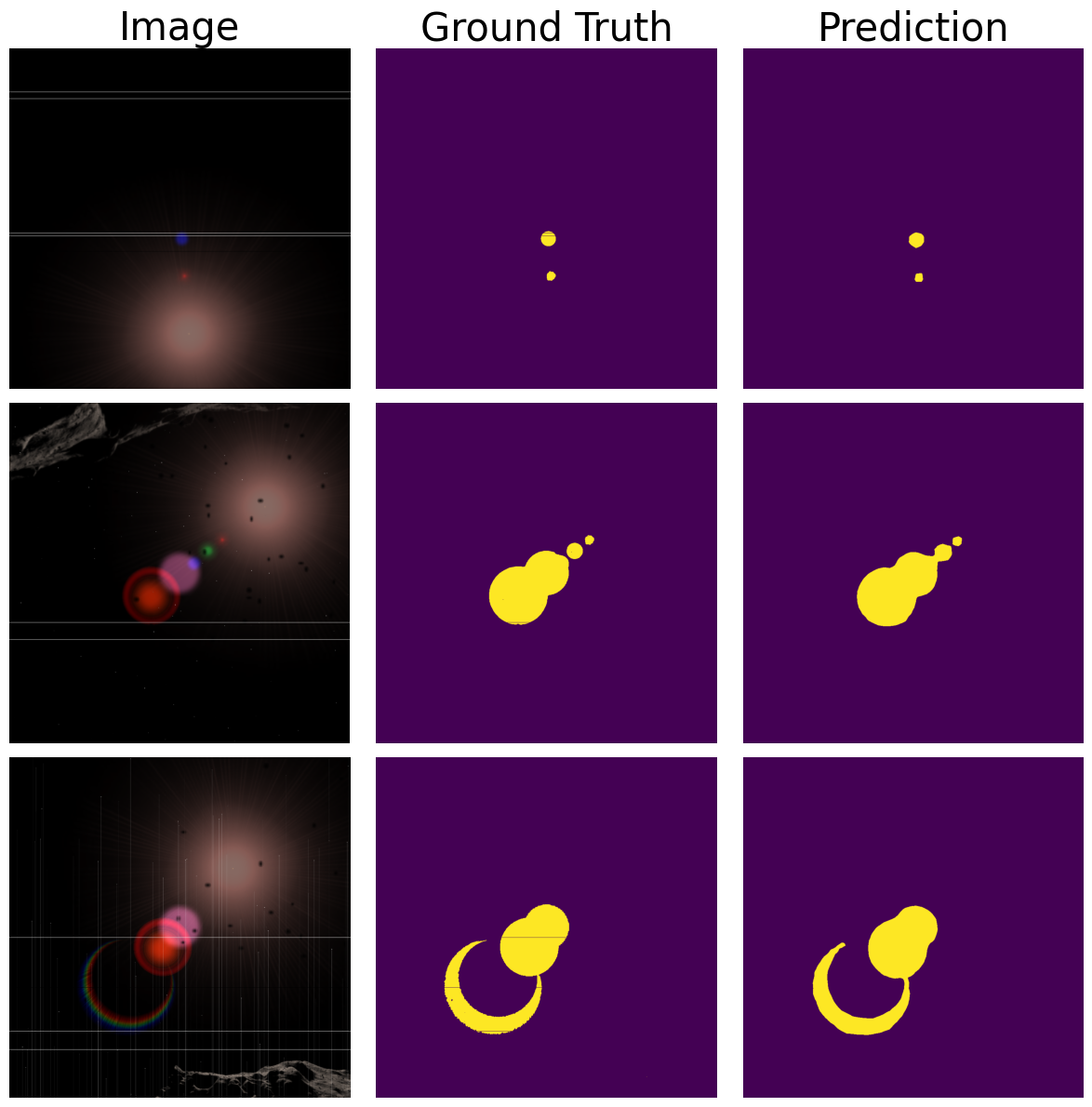}
    \caption{Binary segmentation of straylight. Fine-tuned model inferenced on custom dataset. Training takes 500 epochs, employs Binary Cross-entropy loss and Adam optimizer with $0.0001$ initial learning rate.}
    \label{fig:binclass_sl_ft}
\end{figure}

Visual inspection indicates that the model is indeed capable of segmenting the desired textures, albeit with insufficient accuracy. Furthermore, the Sun itself is segmented as a flare. This outcome suggests the correctness of our approach, as the Sun's rendering in the images shares the same characteristics as the flare rendering, indicating the model is segmenting both based on common learned features.\\
Finally, we resumed training by feeding the model with our custom dataset. The results are presented in Figure \ref{fig:binclass_sl_ft}, showing a consistent improvement in the Sun segmentation and flares' shape learning.

\section{Evaluation and Results} \label{sec:ev_res}
The evaluation of the straylight segmentation model presented in this work will be based on conventional metrics first, then on custom metrics which reflect the system-level impact of such model when integrated onboard. 
\begin{figure}[h!]
    \centering
    \includegraphics[width=\linewidth]{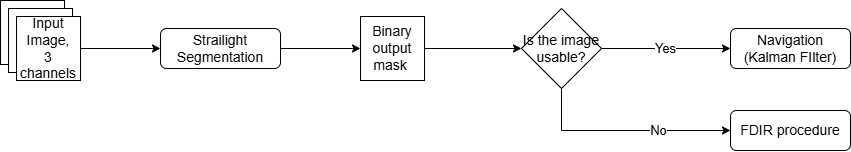}
    \caption{Onboard camera sensor data processing scheme}
    \label{fig:onb_proc}
\end{figure}
In Figure \ref{fig:onb_proc} we provide a scheme of how the output of our segmentation model shall be processed onboard. It is common practice in modern satellites' Guidance, Navigation and Control (GNC) subsystems to include validity checks on sensor data to assess their usability in downstream algorithms. In this case, the output of the model is a binary segmentation mask which is marking the per-pixel validity based on straylight presence. This validity can be processed onboard to first assess whether the image is usable for navigation purposes or an FDIR procedure shall be triggered to recover the sensor functionality. As long as the image is declared usable, it can be employed in the Kalman filter to provide information on the spacecraft state, excluding from the update step of the filter those pixels declared invalid. This processing strategy improves the availability of the navigation pipeline, which would be capable to function in a fail-operational state. Marking unusable images is a significant challenge of this approach, and it is left for future work.

\begin{figure}[h!]
    \centering
    \begin{subfigure}[b]{0.47\textwidth}
         \centering
         \includegraphics[width=\textwidth]{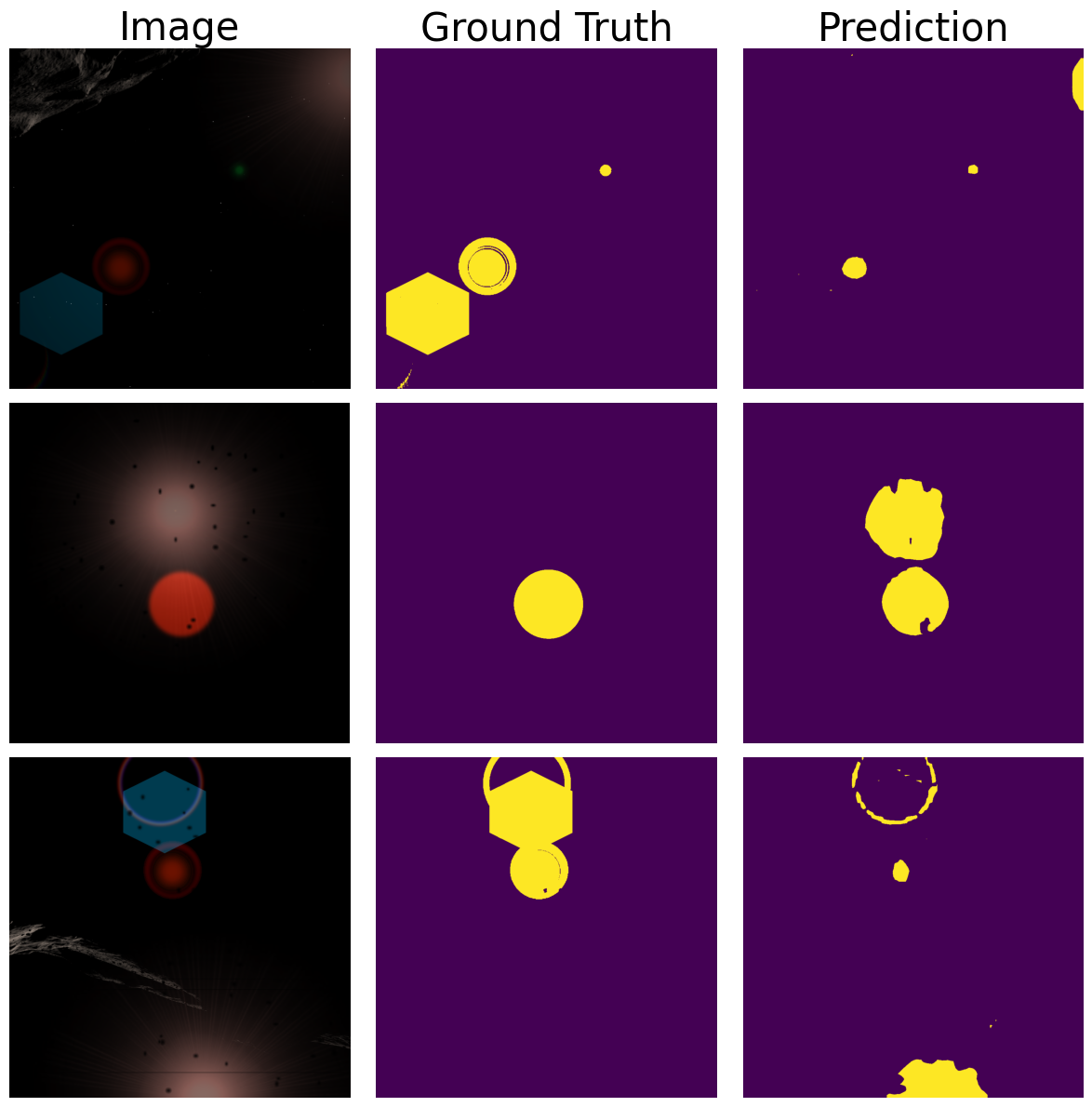}
     \end{subfigure}
     \hfill
     \begin{subfigure}[b]{0.47\textwidth}
         \centering
         \includegraphics[width=\textwidth]{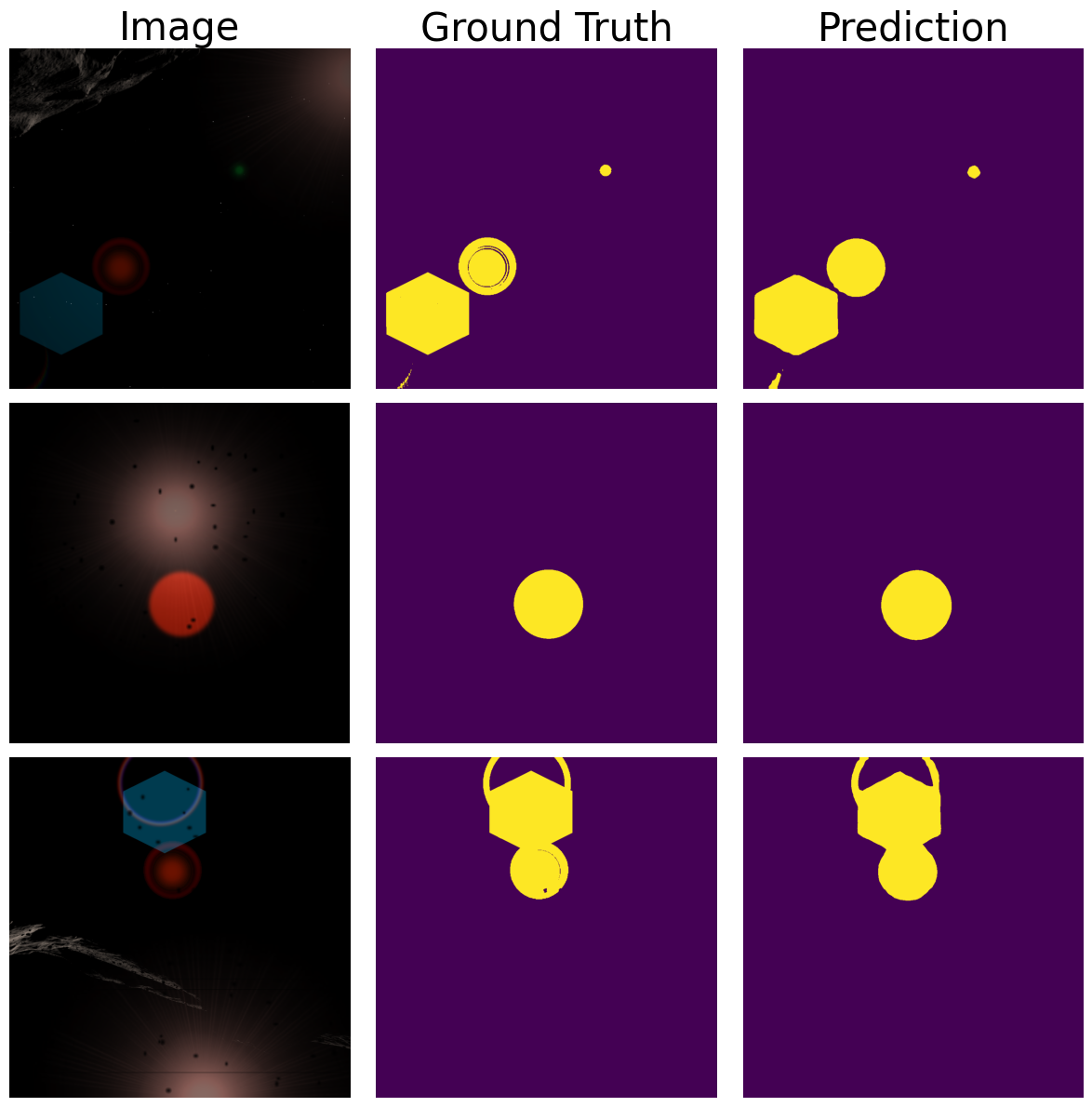}
     \end{subfigure}
      \caption{More segmentation results for the pre-trained (left) and fine-tuned (right) models}
     \label{fig:binclass_sl_both}
\end{figure}

We evaluate the performances of the straylight segmentation model employing conventional metrics first. Table \ref{tab:results_classical} shows the results obtained inferencing the pre-trained and the fine-tuned model on our custom test dataset. In addition, Figure \ref{fig:binclass_sl_both} shows more examples of faulty images inferenced on both models, highlighting the consistent improvement brought by fine-tuning.
\begin{table}[h!]
    \centering
    \begin{tabular}{|c|c|c|c|}
        \hline
         & Precision & Recall & mIoU \\ \hline
        Pre-trained Model & 0.259 & 0.403 & 0.188\\\hline
        Fine-tuned Model & 0.908 & 0.958 & 0.873\\\hline
    \end{tabular}
    \caption{Segmentation results for the Straylight model.}
    \label{tab:results_classical}
\end{table}

The pre-trained model shows degraded performances on the custom dataset with respect to the fine-tuned one due to the biases already identified in Section \ref{subs:dev_proc}. Such a model is capable of grasping shapes of most of the straylight artifacts, although it is systematically segmenting the Sun as a flare. This erroneous behaviour determines a consistently lower Precision before fine-tuning, after which the Sun is correctly segmented instead. Results in Table \ref{tab:results_classical} also show a Recall increase after fine-tuning, as the model prediction globally improves accuracy and better adheres to the flares' actual shape (Figure \ref{fig:binclass_sl_both}). For the same reason, mIoU also raises due to the better coincidence of \textit{intersection} and \textit{union} pixels between prediction and ground truth.

A different point of view on the evaluation of the proposed segmentation model is given by the metrics presented in Table \ref{tab:img_metrics}.
\begin{table}[h!]
    \centering
    \begin{tabular}{|c|c|}
    \hline
        Metric Name & Definition \\ \hline
        Per-artifact Recall (PaR) & \rule{0pt}{3.5ex}$1-\dfrac{\mbox{False Negative Artifacts}}{\mbox{Total Ground Truth Artifacts}}$ \\[1.5ex] \hline
        Per-artifact Precision (PaP) & \rule{0pt}{3.5ex}$1-\dfrac{\mbox{False Positive Artifacts}}{\mbox{Total Predicted Artifacts}}$ \\[1.5ex] \hline
        Per-artifact mIoU (PamIoU) & \rule{0pt}{3.5ex}$\dfrac{\mbox{intersection}}{\mbox{union}}, \mbox{except False Positives and False Negatives}$ \\[1.5ex] \hline
    \end{tabular}
    \caption{Custom metrics for image faults segmentation}
    \label{tab:img_metrics}
\end{table}

The focus of the evaluation through these custom metrics is shifted to an artifact-based paradigm \cite{gallon2025convolutional}, meaning that single flares count as whole objects to the scope of metrics computation. While in conventional segmentation metrics (i.e., Precision, Recall and mIoU) per-pixel discrepancies are considered, our approach focuses on regions of contiguous pixels and relates them to the subtended ground truth artifacts, if existent. Evaluating entire artifacts rather than individual pixels provides a broader and more system-oriented perspective on the implications of the model output, particularly when it is processed in the navigation subsystem. For this subsystem, it is critical to detect and isolate all potentially corrupted image regions, even if the segmentation is not perfectly accurate at pixel level, in order to prevent their influence on subsequent processing stages. While fine-grained pixel accuracy remains desirable, it does not convey whether entire artifacts are missed or incorrectly identified as false positives. The proposed custom metrics are therefore designed to supply to this limitation.

\begin{figure}[h!]
    \centering
    \begin{subfigure}[b]{0.47\textwidth}
         \centering
         \includegraphics[width=\textwidth]{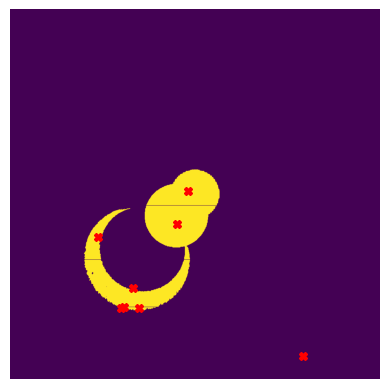}
     \end{subfigure}
     \begin{subfigure}[b]{0.47\textwidth}
         \centering
         \includegraphics[width=\textwidth]{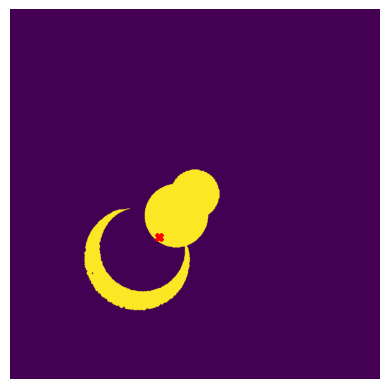}
     \end{subfigure}
      \caption{Connected regions identified by the \textit{regionprops} algorithm are marked by red crosses. Standard ground truth (left) and Gaussian smoothed (right)}
     \label{fig:regionprops}
\end{figure}

Relevant regions in the prediction mask are delineated by means of classical image processing algorithms. Specifically, the \textit{regionprops} algorithm in the \textit{measure} API from the \textit{scikit-image} library is utilized to identify connected regions in both the ground-truth mask and the model prediction with $connectivity=2$. Compatibility between this algorithm and the considered dataset is ensured via a preprocessing step, which in turns allows the evaluation of custom metrics presented in Table \ref{tab:img_metrics}. Figure \ref{fig:regionprops} illustrates the motivation and the effect of the pre-processing step. The dataset's ground truth generation mechanism detailed in \cite{gallon2025addressing} inherently prioritizes the broken line over the flare annotation. This prioritization is an artifact of the sequential rendering process of the fault types listed in \cite{gallon2025addressing}, and would not be present in case of a different faults prioritization case. Consequently, the region corresponding to the presence of a flare is inadvertently perceived as multiple subregions by \textit{regionprops}. This fragmentation introduces a significant bias into the computation of the custom metrics, because it unpredictably increases the number of regions in the ground truth accounting for the computation of e.g., False Negatives and True Positives. Hence, a single contiguous region corresponding to one flare texture in the segmentation output will be compared with many regions in the ground truth, leading to a systematic increase of False Negatives. In other words, too many labels exist for the same flare. The gaps spuriously separating connected regions can have different widths based on the random coincidence of contiguous faults \cite{gallon2025addressing}. We apply a Gaussian smoothing algorithm to fill the empty spaces and restore continuity of the ground truth, ultimately enabling the applicability and representativeness of custom metrics. The Gaussian smoothing is implemented in the \textit{filters} API from \textit{scikit-image}, employing as parameters $sigma=1$ and $truncate=5$. $sigma$ represents the standard deviation of the sampled Gaussian curve, while $truncate$ represents the number of standard deviations where the sampling of the Gaussian is truncated. $truncate \times 2-1$ is the size of the filter matrix which is convolved with the image. We select the $truncate$ value via empirical testing and it is related to the maximum gap between fragmented regions in the whole dataset. 
%The parameter is increased until filling of the gaps and correct identification of ground truth regions is deemed sufficient, based on the trend shown by custom metrics resulting from the variation of the $truncate$ parameter.

It is noteworthy that the Gaussian smoothing algorithm invalidates conventional metrics for the evaluation of the model performances, because the spread of ground truth over an increased number of pixels significantly affects per-pixel computations. In other words, the number of pixels annotated as faulty introduced by the smoothing is not negligible with respect to the overall amount of faulty pixels in the ground truth.
\begin{table}[h!]
    \centering
    \begin{tabular}{|c|c|c|c|}
        \hline
         & PaP & PaR & PamIoU \\ \hline
        Standard Ground Truth & 0.849 & 0.308 & 0.855\\\hline
        Smoothed Ground Truth & 0.832 & 0.991 & 0.783\\\hline
    \end{tabular}
    \caption{Comparison between the proposed custom metrics, before and after Gaussian smoothing, evaluated on the fine-tuned model.}
    \label{tab:custom_metr_gs}
\end{table}

As expected, PaP and PamIoU are shown to be mildly sensitive to the mentioned smoothing. PaP is substantially unchanged because connecting contiguous regions does not modify the true label of the single artifact, hence true and false positive artifacts depend only on the segmentation output, which is untouched. Instead, PamIoU experiences a more consistent degradation due to the increase of newly annotated pixels. Most importantly, Gaussian smoothing consistently improves the number of False Negatives, cancelling out the spurious regions derived from the dataset flaws which are not relatable to any object in the prediction. Figure \ref{fig:regionprops} is exemplary to visualize this effect. The Gaussian smoothing operates on the pixels along the fragmented contours, effectively converting them into True Positives and bridging discontinuities. As a result, the smoothed region in the ground truth becomes continuous and remains associated with the same predicted region as prior to smoothing. Concurrently, disconnected and spurious regions are removed, leading to a cleaner and more coherent segmentation representation.

In conclusion, the described Gaussian smoothing process does not jeopardize performances of the algorithm while correctly acting on the identified bias of the custom metrics.

In Table \ref{tab:results} we present the final results for our custom metrics and correspondent classical segmentation metrics, obtained inferencing the pre-trained and fine-tuned model. 
\begin{table}[h!]
    \centering
    \begin{tabular}{|c|cc|cc|cc|}
        \hline
         & PaP & Precision & PaR & Recall & PamIoU & mIoU \\ \hline
        Pre-trained Model & 0.456 & 0.259 & 0.769 & 0.403 & 0.313 & 0.188\\\hline
        Fine-tuned Model & 0.832 & 0.908 & 0.991 & 0.958 & 0.783 & 0.873\\\hline
    \end{tabular}
    \caption{Results for the Straylight model, as evaluated on custom metrics and conventional segmentation metrics}
    \label{tab:results}
\end{table}

The results highlight a significant difference in the two sets of metrics evaluated on the pre-trained model. Specifically, in the computation of the PaR/Recall metrics, textures blending after the Gaussian smoothing relates different and poorly-segmented textures to the same ground truth one, originating the discrepancy. The low PamIoU highlights that the model is not performing well despite the high PaR value obtained, signalling that the overlap between prediction and ground truth is poor. This idea proves our point stating that the three metrics are equally important to a comprehensive model evaluation. 
Furthermore, the per-artifact evaluation paradigm is important from the system-level point of view, because it is paramount to detect the faulty artifact, more than segmenting them at pixel-level accuracy. Indeed, undetected flare textures would be processed onboard as regular pixels, leading to misunderstanding of the image information (e.g. landmarks, stars, etc.) and to wrong state estimation in the Kalman filter (see Figure \ref{fig:onb_proc}). This idea translates in an increased focus on PaR over PaP during model tuning, meaning on the reduction of false negative artifacts. Such a focus cannot be reached employing conventional metrics as optimization targets. Far less critical but still significant is the value of PaP, which is associated to false positive predicted artifacts. These artifacts would lead to neglect valid pixels and subsequently reduce available information to the Kalman filter for estimation refinement (see Figure \ref{fig:onb_proc}). However, the decreased criticality stems from the fact the the propagation stage in the filter can supply to the lack of measurements, although for a limited amount of time. In this sense, requirements on other sensors responsible of the propagation (e.g., gyroscopes) can be made more stringent to accommodate faults segmentation algorithms with a higher PaR to the cost of more false positives.

\section{Conclusion and Future Work} \label{sec:concl_futwork}

This work presents a novel approach for the onboard segmentation of faults in camera sensors for space applications. The analysis specifically focuses on straylight artifacts (i.e., solar flares) arising when the Sun is in proximity of the FoV. The proposed deep learning model targets robustness in flare segmentation across diverse textures and backgrounds by leveraging pre-training on a realistic flares dataset followed by fine-tuning using our custom dataset derived from our previous work on space camera sensors faults. The model architecture is explicitly designed targeting onboard-compatibility on space-qualified hardware, addressing the critical constraints of memory and computational power. Furthermore, we propose an integration framework for incorporating the model into a functional navigation pipeline. This framework defines the interfaces between the model and navigation algorithms, facilitating the communication of detected faults. To support this task, we introduce a set of custom metrics that pivot the evaluation from pixel-wise accuracy to artifact-centric performance. This shift reflects a prioritization of global artifact detection over precise boundary segmentation, aligning the evaluation with the operational requirements of the system.

Future research shall focus on validating the model's robustness against domain shift by transitioning from synthetic to real-world mission imagery, counteracting the inherently present domain shift. The shift mainly derive from injected flare textures and simulated backgrounds, thus real images shall mainly supply to these two factors. Subsequent efforts shall prioritize the hardware-in-the-loop deployment on space-qualified FPGAs to verify real-time processing capabilities within strict computational constraints. Finally, future work shall refine the integration architecture, where the interface between AI predictions and navigation algorithms shall be functionally demonstrated. Evaluation shall specifically quantify system availability and resilience in the fail-operational state, establishing efficacy of the model in maintaining mission continuity during sensor-level anomalies.

%%       \cite{<label>} ==> [1]

\bibliographystyle{elsarticle-num} 
\bibliography{sample}


\end{document}